\documentclass[conference]{IEEEtran}
\IEEEoverridecommandlockouts
\usepackage{cite}
\usepackage{amsmath,amssymb,amsfonts}
\usepackage{algorithmic}
\usepackage{graphicx}
\usepackage{textcomp}
\usepackage{xcolor}
\def\BibTeX{{\rm B\kern-.05em{\sc i\kern-.025em b}\kern-.08em
    T\kern-.1667em\lower.7ex\hbox{E}\kern-.125emX}}

\usepackage{flushend}
\usepackage[caption=false]{subfig}
\usepackage{siunitx}
\renewcommand{\vec}[1]{\mathbf{#1}}
\usepackage[hyphens]{url}
\usepackage{hyperref}
\usepackage{mathtools}
\begin{document}
\title{Autonomous Golf Putting with\\ Data-Driven and Physics-Based Methods}

\makeatletter
\def\ps@IEEEtitlepagestyle{%
	\def\@oddfoot{\mycopyrightnotice}%
	\def\@oddhead{\hbox{}\@IEEEheaderstyle\leftmark\hfil\thepage}\relax
	\def\@evenhead{\@IEEEheaderstyle\thepage\hfil\leftmark\hbox{}}\relax
	\def\@evenfoot{}%
}
\def\mycopyrightnotice{%
	\begin{minipage}{\textwidth}
		\centering \scriptsize
		© 2022 IEEE. Personal use of this material is permitted. Permission from IEEE must be obtained for all other uses, in any current or future media, including	reprinting/republishing this material for advertising or promotional purposes, creating new	collective works, for resale or redistribution to servers or lists, or reuse of any copyrighted component of this work in other works.
	\end{minipage}
}
\makeatother

%\thanks{© 2022 IEEE. Personal use of this material is permitted. Permission from IEEE must be obtained for all other uses, in any current or future media, including	reprinting/republishing this material for advertising or promotional purposes, creating new	collective works, for resale or redistribution to servers or lists, or reuse of any copyrighted component of this work in other works.}

\author{\IEEEauthorblockN{Annika Junker\IEEEauthorrefmark{1},
		Niklas Fittkau\IEEEauthorrefmark{4},
		Julia Timmermann\IEEEauthorrefmark{3},
		Ansgar Trächtler\IEEEauthorrefmark{2}}
\IEEEauthorblockA{\textit{\hspace{0pt}Heinz Nixdorf Institute,}
\textit{Paderborn University,}
Paderborn, Germany\\
Email: \IEEEauthorrefmark{1}annika.junker@hni.upb.de, 
\IEEEauthorrefmark{4}niklas.fittkau@hni.upb.de,\\ 
\IEEEauthorrefmark{3}julia.timmermann@hni.upb.de, 
\IEEEauthorrefmark{2}ansgar.traechtler@hni.upb.de}
}

\maketitle

\begin{abstract}
We are developing a self-learning mechatronic golf robot using combined data-driven and physics-based methods, to have the robot autonomously learn to putt the ball from an arbitrary point on the green. Apart from the mechatronic control design of the robot, this task is accomplished by a camera system with image recognition and a neural network for predicting the stroke velocity vector required for a successful hole-in-one. To minimize the number of time-consuming interactions with the real system, the neural network is pretrained by evaluating basic physical laws on a model, which approximates the golf ball dynamics on the green surface in a data-driven manner. Thus, we demonstrate the synergetic combination of data-driven and physics-based methods on the golf robot as a mechatronic example system.
\end{abstract}

\begin{IEEEkeywords}
hybrid modeling, nonlinear control, autonomous systems, robotics, machine learning
\end{IEEEkeywords}

\section{Introduction}\label{sec:introduction}
With the aid of autonomous robots, the everyday life of many people should be made easier in the near future, e.g., by supporting work in the care of elderly or physically impaired people. For this, a prudent action of the autonomous robot is essential. We have included these aspects in our research work and want to develop them further in the context of an autonomous golf robot as an illustrative example. For an autonomous golf game in the area of the hole, some challenges have to be overcome, e.g., sophisticated environment recognition and precise actuation. Even for skilled humans, putting is not always successful. To be able to meet these requirements, in this paper we use both powerful data-driven methods and established physics-based methods from the control engineering context. A hybrid approach is extremely beneficial to optimally utilize the advantages from both areas.  

In the field of golf sports, there are some robots, which perform a wide variety of tasks. One area is the support of players with regard to their stroke execution. For example, the robot in \cite{Rob17} optimizes the swing by directly guiding the golfer's arm. Another very common application of golf robots is testing equipment. \cite{Bet10,CHV16,Gol03,The98,Miy,Miyb,Miyc} are examples of robots to test golf clubs and balls using many different strokes, where the stationary constructions mostly consist of a rotating robotic arm. The third area of application imitates human strokes. The robot \textit{ROB-OT} \cite{The18} plays a complete golf game and thus serves as a demonstrative training for golfers, but also for entertainment purposes. Although it moves around the entire green, it requires a golf-experienced operator during play.

\begin{figure}[t!]
	\centering
	\includegraphics[width=6.5cm]{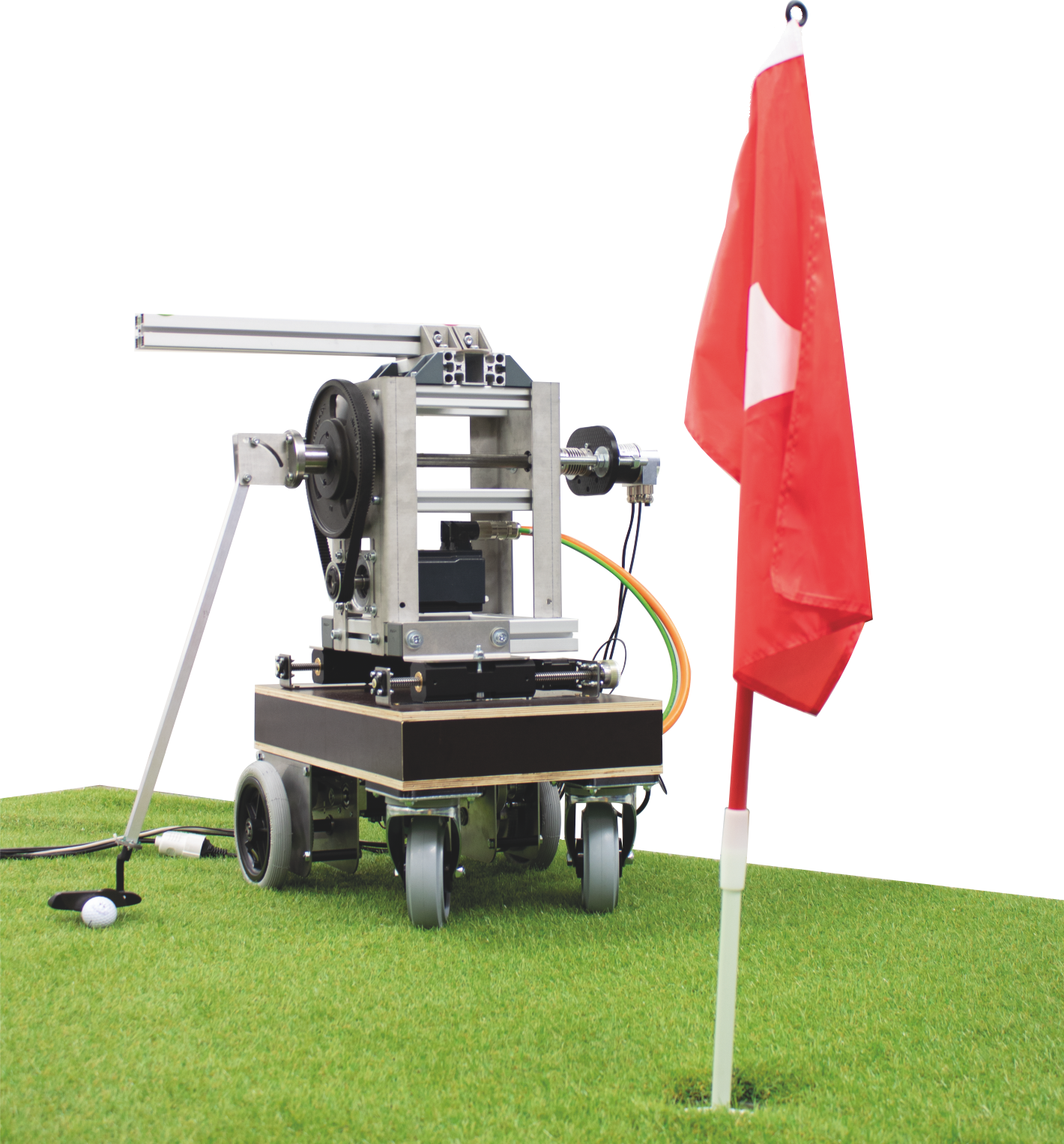}
	\caption{\textit{Golfi} is a self-learning golf robot, which is able to putt autonomously and serves as a demonstrator for data-driven methods in control engineering.}
	\label{fig:golfrobot}
\end{figure}
\begin{figure}[t!]
	\centering
	\advance\leftskip-0.18cm
	\includegraphics[width=\linewidth]{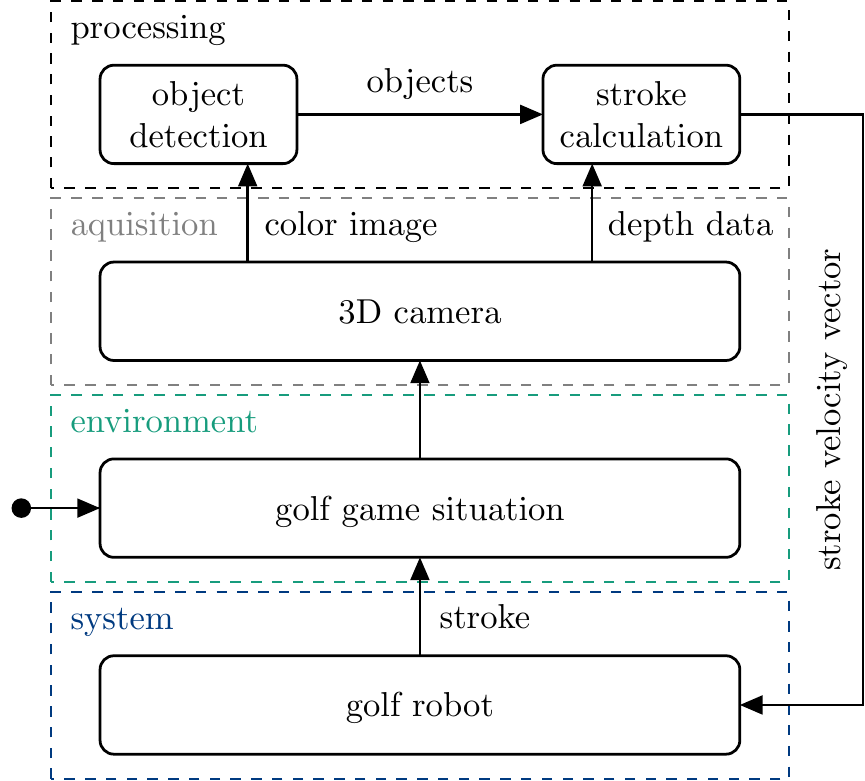}
	\caption{System architecture of the golf robot: The overall problem is decomposed into separate sub-problems, some of which are data-driven, others physics-based, and still others hybrid. The starting point of the process is a given golf game situation.}
	\label{fig:overview}
\end{figure}

Our golf robot \textit{Golfi}, shown in Fig.~\ref{fig:golfrobot}, is characterized by its aim to putt completely autonomously. This means that it is supposed to hit the ball into the hole from an arbitrary initial position on an unknown green by a single stroke, without the ball flying in the air. We use a combination of classical control engineering and data-driven techniques to solve this task. The positioning and stroke devices can be physically modeled in a simple way, so that classical engineering methods are used for the controller design. In contrast, analyzing the game situation and determining an optimal stroke direction and velocity are extremely challenging problems that cannot be easily solved using basic physical laws. We structure the task of autonomous golf putting into separate sub-problems, as shown in Fig.~\ref{fig:overview}. The complexity increases from bottom to top and with it the proportion of data-driven approaches. At the bottom level is the mechatronic system of the golf robot, which moves on the green and executes strokes with a given velocity vector. The positions of Golfi, the ball, and hole as well as the shape of the green surface define the golf play situation and are captured by a 3D camera. At the top level, we use data-driven methods to detect the objects and a synergetic combination of data-driven and physics-based methods to calculate the required stroke velocity vector to putt the ball into the hole. 
This strategy is to first pretrain a neural network using training strokes generated simulatively using a physics-based ball dynamics model once for a given green. Afterwards, we aim to retrain the neural network using training strokes on the real system. This procedure significantly reduces the number of time-consuming interactions with the real system and provides a superior result by combining data-driven and physics-based methods in a goal-oriented and meaningful way.

The paper is structured as follows: First, we describe the mechatronic design of the robot, cf.~Sec.~\ref{sec:mechatronic_design}. Then we explain the computer vision system in Sec.~\ref{sec:computer_vision} and our hybrid strategy to determine an optimal stroke velocity vector in Sec.~\ref{sec:optimal_stroke_velocity_vector}. Sec.~\ref{sec:results} presents the results and Sec.~\ref{sec:conclusion} concludes with a summary and outlook. 

\textbf{Notation:} All vectors and matrices are printed in bold. The index at the bottom left indicates the coordinate system in which the vectors are described, e.g., $\prescript{}{\mathrm{I}}{\vec{v}_s}$ denotes the stroke velocity vector, described in the coordinate system I. $\lVert\prescript{}{\mathrm{I}}{\vec{v}_s}\rVert$ denotes the length of the vector $\prescript{}{\mathrm{I}}{\vec{v}_s}$, given by the Euclidian norm.
\section{Mechatronic design of the golf robot}\label{sec:mechatronic_design}
The mechatronic system of the golf robot needs to realize a given stroke velocity vector, which means to hit the ball in a specific direction at a specific velocity. This task is divided into two sub-tasks:
\begin{enumerate}
	\item the positioning device must place Golfi so that the club is next to the ball in a specific direction and
	\item the stroke device must then hit the ball so that it starts rolling with a specific initial velocity,
\end{enumerate}
which are described in Sec.~\ref{subsec:positioning_device} and  \ref{subsec:stroke_device}, respectively. 
\subsection{Positioning device}\label{subsec:positioning_device}
\begin{figure}[t!]
	\centering
	\includegraphics[]{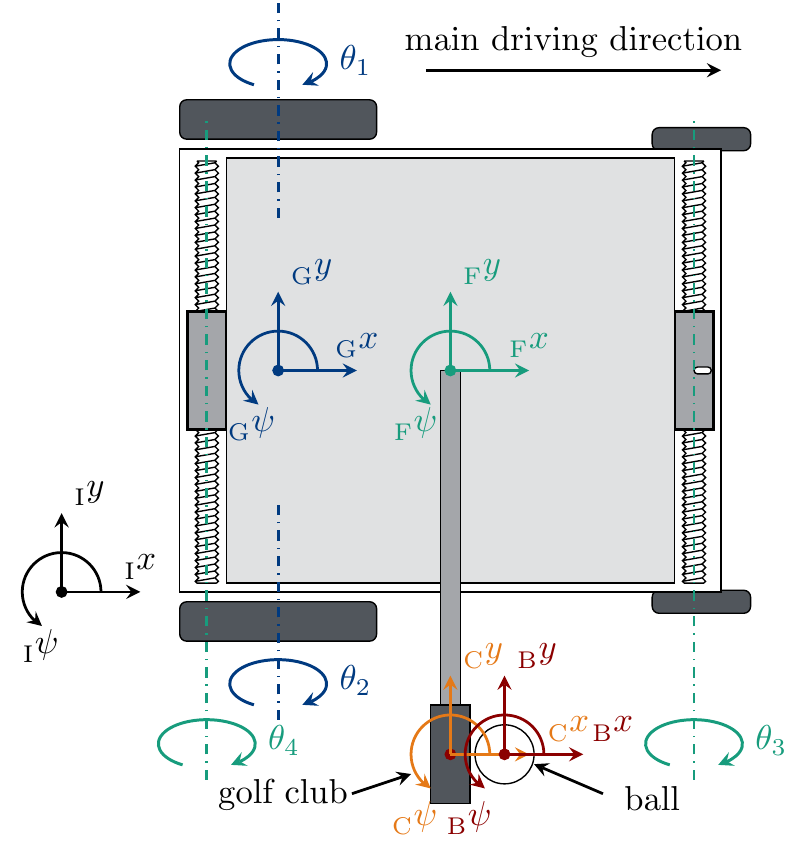}
	\caption{The positioning device, here in top view, consists of a drive unit and a fine traversing unit. The drive unit consists of two rear wheels, which are driven individually (in blue) with torques $\theta_1$ and $\theta_2$, with the weight of the robot supported by the two swivel castors at the front. It allows the robot to rotate by the angle $\prescript{}{\mathrm{G}}{\psi}$ and execute movement along the $\prescript{}{\mathrm{G}}{x}$-axis. The fine traversing unit has a slotted hole on the right cart, which allows the robot to rotate by small angles $\prescript{}{\mathrm{F}}{\psi}$ by controlling the torques $\theta_3$ and $\theta_4$ in opposite directions of movement as well as movement along the $\prescript{}{\mathrm{F}}{y}$-axis.}
	\label{fig:positioning_mechanism}
\end{figure}

The device for positioning the robot consists of a drive unit and a fine traversing unit, which is shown in Fig.~\ref{fig:positioning_mechanism}. Since we are looking at the green from a bird's eye view, the positioning problem simplifies into a plane problem, where only the $x$-axis, $y$-axis, and planar rotation are sufficient to describe the positioning. The inertial coordinate system I, which corresponds to the camera coordinate system, is located approximately in the center of the green. The body-fixed coordinate system G is located between the two rear wheels and oriented with its $\prescript{}{\mathrm{G}}{x}$-axis pointing forward and its $\prescript{}{\mathrm{G}}{y}$-axis pointing to the left. 

The drive unit, which comprises a chassis with 2~driven JMC servo motors controlled by an Arduino and 2~freely rotating wheels, realizes a translation along the $\prescript{}{\mathrm{G}}{x}$-axis and a rotation by the angle $\prescript{}{\mathrm{G}}{\psi}$. Although Golfi can theoretically adopt any pose with it, the translational movement along the $\prescript{}{\mathrm{G}}{y}$-axis is difficult to realize so we additionally aim to use a fine traversing unit described by the coordinate system F, which is located between the chassis and the stroke device and is based on two spindles driven by Joy-IT stepper motors with Leadshine DM 542 drivers. This fine traversing unit enables small but far more precise movement of the robot as the influence of the traction to the ground is removed here. Within the coordinate system F, translational movement along the $\prescript{}{\mathrm{F}}{y}$-axis and rotation by the angle $\prescript{}{\mathrm{F}}{\psi}$ in the range of $\pm\SI{18}{\degree}$ is possible. The wheels and spindles can be directly given translational travel commands. For the positioning units, we assume the travel commands to be ideally realized since we use stepper motors. We furthermore introduce the club coordinate system C, which is located at the center of the club as well as the ball coordinate system B, which is located at the center of the ball and has the same orientation as I. The final goal for the positioning is to align the $\prescript{}{\mathrm{C}}{x}$-axis with the direction of the desired stroke velocity vector $\prescript{}{\mathrm{I}}{\vec{v}_s}$ and to place the origin of the coordinate system C with an offset of $\SI{3}{\centi\meter}$ in negative $\prescript{}{\mathrm{I}}{\vec{v}_s}$ direction starting from the origin of B. In our first tests presented here, we currently only use the driving unit and not the fine traversing unit, which nevertheless provided sufficient results. 

The pose of the golf club is denoted as $\prescript{}{\mathrm{I}}{\vec{g}}_\mathrm{C}=\left[\prescript{}{\mathrm{I}}{x}_\mathrm{C},\prescript{}{\mathrm{I}}{y}_\mathrm{C},\prescript{}{\mathrm{I}}{\psi}_\mathrm{C}\right]^\top$. After positioning, the goal is that the end pose of the robot's club ${\prescript{}{\mathrm{I}}{\vec{g}_{\mathrm{C},e}}}$ corresponds to the desired target pose  ${\prescript{}{\mathrm{I}}{\vec{g}_{\mathrm{C},d}}}$, depending on the ball position and the stroke velocity vector. The required control signals for the stepper motors are calculated by minimizing the following optimal control objective
\begin{equation}
	\begin{aligned}
		J_p(\vec{\Theta})=&(\prescript{}{\mathrm{I}}{\vec{g}_{\mathrm{C},d}}-\prescript{}{\mathrm{I}}{\vec{g}_{\mathrm{C},e}})^\top\vec{Q}_p(\prescript{}{\mathrm{I}}{\vec{g}_{\mathrm{C},d}}-\prescript{}{\mathrm{I}}{\vec{g}_{\mathrm{C},e}})
		\\&+\sum_{i=1}^{N}\vec{\Theta}^\top(i)\vec{R}_p\vec{\Theta}(i),
	\end{aligned}
\end{equation}
where 
\begin{equation}
	\vec{\Theta}=\left[
		\vec{\Theta}(i),\cdots,\vec{\Theta}(N)
	\right]\text{ with }\vec{\Theta}(i)=\begin{bmatrix}
		\theta_1(i)\\\theta_2(i)
	\end{bmatrix}
\end{equation} 
is the sequence of the control signals $\theta_1$, $\theta_2$ and $\vec{Q}_p$, $\vec{R}_p$ are weighting matrices. The end pose  $\prescript{}{\mathrm{I}}{\vec{g}_{\mathrm{C},e}}$ of the robot is based on the model-based kinematics of the positioning device. The optimization problem is solved in MATLAB using particle swarm optimization. 

\subsection{Stroke device}\label{subsec:stroke_device}
The stroke device consists of two gear shafts connected with a toothed belt drive. The drive (Beckhoff AM8042) is located on the lower gear shaft, while the golf club is mounted on the upper gear shaft. A simplified nonlinear model combines the masses into a single body with torque $u$ as control input and an ideal gear ratio of $4$. The differential equations with parameters shown in Table \ref{tab:stroke_parameters} can be described by the following:
\begin{equation}
	\begin{aligned}
		\dot{x}_1&=x_2,\\
		\dot{x}_2&=\tfrac{-m_cga\sin{x_1}-M_d(\vec{x})+4u}{J},
	\end{aligned} 
\end{equation}
where the state vector $\vec{x}=\left[
	x_1,x_2
\right]^\top=\left[
	\varphi,\dot{\varphi}
\right]^\top$ contains the angle and angular velocity of the golf club and the nonlinear damping dissipation torque 
\begin{equation}
	M_d(\vec{x})=dx_2+r\mu_c\mathrm{sgn}{x_2}\lvert m_Cx_2^2 a+m_Cg\cos{x_1}\rvert
\end{equation}
combines viscous and sliding friction. The angle $\varphi$ is measured directly, so is $y=x_1$, requiring a state observer for the estimation of $x_2$.
\begin{table}[t!]
	\caption{Physical parameters of the stroke device.}
	\label{tab:stroke_parameters}
	\begin{center}
		\begin{tabular}{c p{4.2cm} p{1.9cm}}
			%			\hline
			symbol&physical parameter&value\\
			\hline
			$m_c$& mass of the golf club&  \SI{0.5241}{\kilo\gram}\\ 
			$J$& inertia of the golf club&  \SI{0.1445}{\kilo\gram\per\meter^2}\\
			$g$& gravity constant& \SI{9.81}{\meter\per\second^2}\\
			$a$& length from the axis of rotation to the center of mass of the golf club& \SI{0.4702}{\meter}\\
			$d$& dynamic friction constant& \SI{0.0132}{\kilo\gram\meter^2\per\second}\\
			$r$& length from the axis of rotation to the friction point& \SI{0.0245}{\meter}\\
			$\mu_c$& static friction constant& $1.5136$\\
			$h$& length from the axis of rotation to the hitting point on the club iron &  \SI{0.6}{\meter}\\ 
			$T_{l}$& duration for lunge and reset& \SI{0.35}{\second}\\ 
			$-\varphi_{l}=\varphi_r$& angle for lunge and reset&  \SI{0.9}{\radian}\\ 
			\hline
		\end{tabular}
	\end{center}
\end{table}

\begin{figure}[t!]
	\centering
	\includegraphics[scale=0.9]{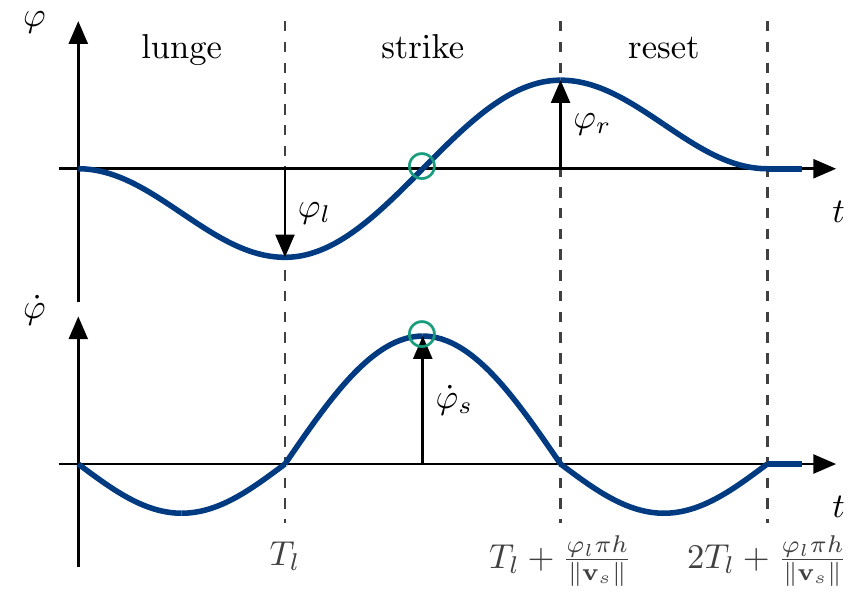}
	\caption{The reference trajectories for $\varphi$ and $\dot{\varphi}$ consist of three phases (lunge, strike, reset) and are thus described by a piecewise-defined function. $\dot{\varphi}_s$ is the desired stroke velocity, whereas $\varphi_l$, $\varphi_r$ and $T_l$ are the lunge and reset angles and lunge duration, respectively. It can be seen that the desired stroke velocity is specified at the point where the angle is zero, which is marked by green circles.}
	\label{fig:reference_trajectory}
\end{figure}

The desired translational stroke velocity at the club head is given by the absolute value of the desired stroke velocity vector $\lVert\vec{v}_s\rVert$, which is to be realized, when the club points vertically downwards, i.e., at a club angle of ${\varphi=\SI{0}{\radian}}$. The controlled rotational velocity of the stroke device is thus \begin{equation}
	\dot{\varphi}_s=\tfrac{\lVert\vec{v}_s\rVert}{h}
\end{equation} with $h$ being length from the axis of rotation to the hitting point on the club iron. Following this idea, reference trajectories $\vec{w}=\left[
	\varphi,\dot{\varphi}
\right]^\top$ are specified, which consist of the three phases \textit{lunge, strike, reset}, see Fig.~\ref{fig:reference_trajectory} and Appendix~\ref{app}.
 
\begin{figure}[t!]
	\centering
	\includegraphics[width=\linewidth, keepaspectratio]{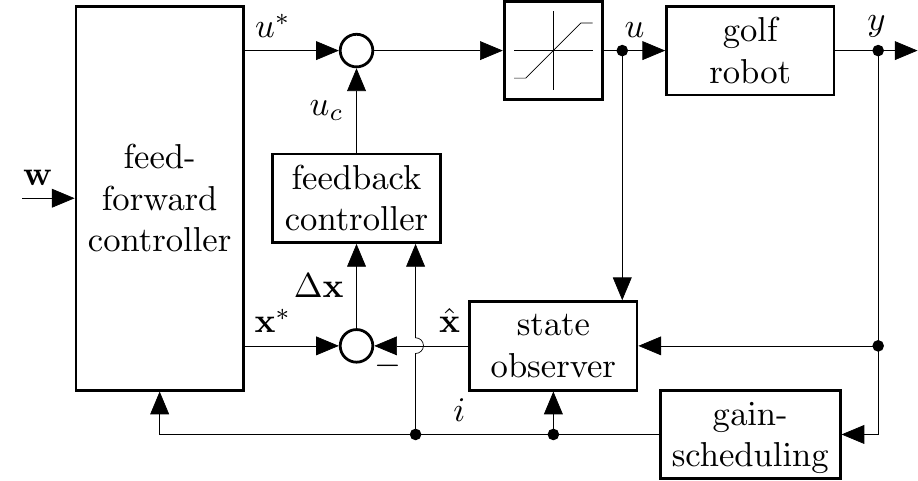}
	\caption{The control strategy for the stroke device is based on a two-degree-of-freedom structure with gain-scheduling, where the scheduling variable $i$ determines both the feedforward and feedback controller as well as the state observer.} 
	\label{fig:control_scheme}
\end{figure}

The feedback control is realized using a gain-scheduling approach with a two-degree-of-freedom structure, where the full state vector $\vec{x}$ is estimated using a state observer, as shown in Fig.~\ref{fig:control_scheme}. For this purpose, $N$ different operating points ${\varphi}_{R_i}$ with ${\varphi_{R_i}\in\left[-\pi,\pi\right]}$ with an increment of $\SI{0.01}{\radian}$ are defined. During operation, the control strategy then switches between the different linearized systems by 
\begin{align}
	i & =\mathrm{arg}~ \underset{i}{\min} \left|x_1-\varphi_{Ri}\right|,
\end{align} 
where $i$ is the scheduling variable.
For the controller and observer design, the nonlinear system dynamics is linearized at each operating point $\varphi_{R_i}$, yielding
\begin{equation}
	\begin{aligned}
			\dot{\vec{x}}&=\vec{A}_{R_i}\vec{x}+\vec{b}u,\\
		y&=\vec{c}^\top\vec{x}
	\end{aligned}
\end{equation}
with
\begin{equation}
	\begin{aligned}
		\vec{A}_{R_i} &= \begin{bmatrix} 0 & 1 \\ -\tfrac{m_cga}{J}\cos(\varphi_{Ri}) & -\tfrac{d}{J} \end{bmatrix},\\
		\vec{b}  &= \left[ 0 , \tfrac{4}{J} \right]^\top,	\vec{c}^\top = \left[ 1, 0 \right].
	\end{aligned} 
\end{equation} 
The linear-quadratic-regulator (LQR) is calculated for
\begin{equation}
	\vec{Q}  = \begin{bmatrix} 5 & 0 \\ 0 & 1 \end{bmatrix},R=1
\end{equation} to determine the linear feedback control law $u_c=-\vec{k}_{R_i}\vec{x}$. The feedforward control matrices are given by
\begin{align}
	\vec{F}_x  = \begin{bmatrix} 1 & 0 \\ 0 & 1 \end{bmatrix}, 
	\vec{F}_{u_{R_i}}  = -\frac{1}{b_{21}}\left[ A_{R_{i_{21}}}, A_{R_{i_{22}}} \right],
\end{align}
defining the reference signals \begin{equation}
	{\vec{x}^*=\vec{F}_x\vec{w}}, {u^*=\vec{F}_{u_{R_i}}\vec{w}}.
\end{equation} Note here that $\vec{F}_{u_{R_i}}$ also depends on the scheduling variable~$i$ and $A_{R_{i_{21}}}$, $A_{R_{i_{21}}}$ and $b_{21}$ denote the matrix elements of $\vec{A}_{R_i}$ and $\vec{b}$, respectively. The resulting equation for the control signal is thus given by
\begin{equation}
	u=u^*+u_c,
\end{equation}
in consideration of the control limits by a saturation.

The design of the Luenberger state observer with
\begin{align}
	\dot{\hat{\vec{x}}} & = (\vec{A}_{R_i}-\vec{L}_{R_i}\vec{b})\hat{\vec{x}}+\vec{b}u+\vec{L}_{R_i}y
\end{align}
is performed analogously for the scheduled system, where the eigenvalues of the observer matrix $\vec{L}_{R_i}$ are twice as far to the left as the eigenvalues of the closed control loop \cite{Oga02}. 

\begin{figure}[t!]
	\centering
	\includegraphics[width=\linewidth, keepaspectratio]{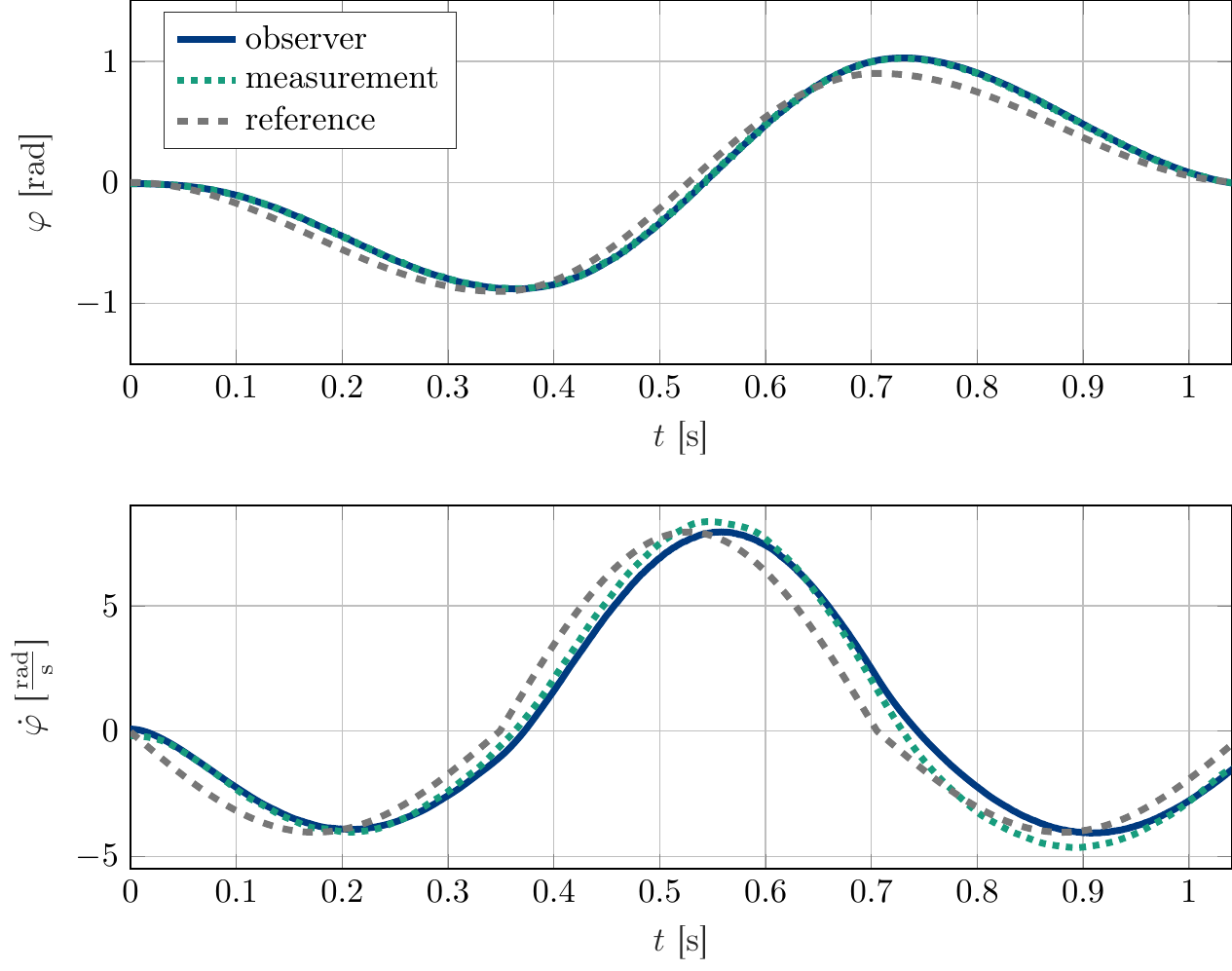}
	\caption{Example for the result of a controlled stroke by the golfrobot. The desired stroke velocity of $\dot{\varphi}_s=\SI{8}{\radian\per\second}$ is to be realized at a club angle of $\varphi=\SI{0}{\radian}$.} 
	\label{fig:stroke_example}
\end{figure}

The resulting control performance is shown exemplarily for a reference stroke velocity of $\dot{\varphi}_s=\SI{8}{\radian\per\second}$, cf.~Fig.~\ref{fig:stroke_example}.
\section{Computer vision}\label{sec:computer_vision}
For a successful golf game, it is necessary to obtain information about the game situation with sensor technology, where we pragmatically use a Microsoft Kinect v2 3D camera mounted on the ceiling resulting in a bird's eye view. It provides an RGB picture and a depth picture and thus is suitable to solve the following two problems:
\begin{enumerate}
	\item object detection and
	\item surface approximation of the green.
\end{enumerate}
The following describes the required procedures and algorithms to detect the poses of the objects such as the ball and the robot, see.~Sec.~\ref{subsec:object_detection} and the necessity of terrain information and its implementation, see~Sec.~\ref{subsec:terrain_information}.
\subsection{Object detection}\label{subsec:object_detection}
The edges of the green and the position of the hole are selected as fixed values. The ball position is detected by a pretrained deep convolutional neural network called Faster R-CNN \cite{RHGS15} in MATLAB, which has been trained with 50 training images. The robot has two contrasting colored circles in pink and light green, cf.~Fig.~\ref{fig:example_situation}, mounted on the top to determine its pose from the bird's eye view. These circles have known positions in relation to the club head coordinate system C, cf.~Fig.~\ref{fig:positioning_mechanism}. The detection of the colored circles is based on scanning the entire picture for the desired RGB values describing pink and light green. Due to a variance of the red, green, and blue channel values of $\pm6$ each, several pixels are always found regardless of light differences. By forming the median of all found ones, one pixel is chosen for each that lies centrally on the color circles, even if pixels were found that lie far away from where the colored circles are.

The positions of the objects, which are detected as pixels in the RGB matrix, are transferred by the point cloud matrix of the depth picture into the inertial coordinate system I, which corresponds to the camera coordinate system. 
\subsection{Terrain information}\label{subsec:terrain_information}
For terrain information, we use an approximated surface in MATLAB. We obtain the 3D data by the point clouds provided by the camera. A differentiable model of the green surface in $\prescript{}{\mathrm{I}}{x}$- and $\prescript{}{\mathrm{I}}{y}$-direction is needed to calculate the slope forces in the differential equation of the golf ball, cf.~Sec.~\ref{subsec:model_based_training_data}. Before setting a specific golf game situation, the green is once captured without objects and approximated. Afterwards, the approximated surface is used for any shots from different positions until the green surface changes.
\section{Optimal stroke velocity vector}\label{sec:optimal_stroke_velocity_vector}
\begin{figure}[h!]
	\centering
	\includegraphics[]{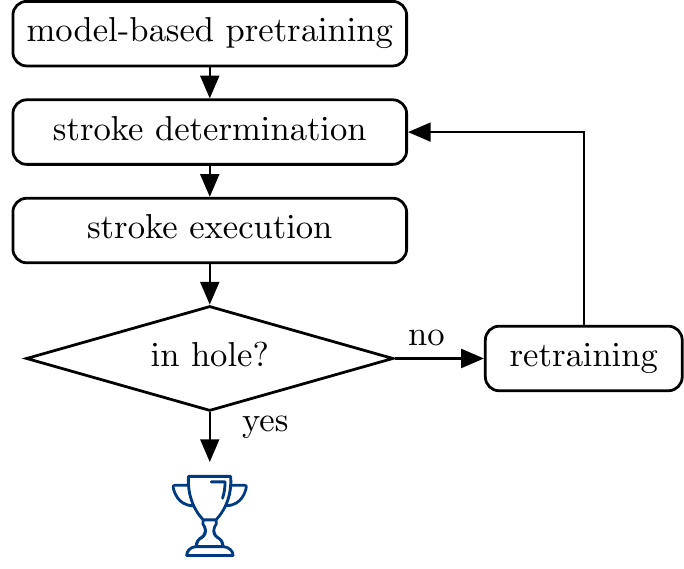}
	\caption{Our training strategy for the autonomous golf game consists of several steps. If the hole is not hit immediately, Golfi can be ordered to take the situation as a new starting point for another stroke. Furthermore, the failed shot may be used to retrain the strategy, that determines the optimal stroke velocity vector.}
	\label{fig:pretraining_strategy}
\end{figure}

The strategy to determine an optimal stroke velocity vector for successful putting, see Fig.~\ref{fig:pretraining_strategy}, is based on several steps:
\begin{enumerate}
	\item Pretraining of a neural network, which represents the golf ball dynamics for a given green surface by simulatively generated training strokes, cf.~Sec.~\ref{subsec:model_based_training_data}.
	\item Determination of an optimal stroke velocity vector based on the neural network, so that the ball hits the hole, cf.~Sec.~\ref{subsec:determine_stroke_velocity_vector}.
	\item Execution of the determined stroke on the real golf green. If the ball does not roll into the hole, Golfi can be ordered to take the situation as a new starting point for another stroke by again using computer vision to determine a new stroke velocity vector, moving to the ball, and attempting to hit the ball into the hole. 
	\item It is feasible to use a failed stroke for retraining the neural network. We have not implemented the retraining process yet but can imagine that after a failed stroke, it may be fed back as another training data point for the strategy.
\end{enumerate}
\subsection{Generate model-based training data}\label{subsec:model_based_training_data}
The model-based pretraining is based on simulatively generated training strokes by evaluating physics-based differential equations of motion. The golf ball dynamics with the parameters given by Table~\ref{tab:ball_parameters} is derived from the initial velocity of the golf ball, the surface of the green, and the rolling resistance of the turf. 

\begin{figure}[t!]
	\centering
	\subfloat[inclined plane, here schematically shown only in two dimensions\label{fig:inclined_plane1}]{\includegraphics[]{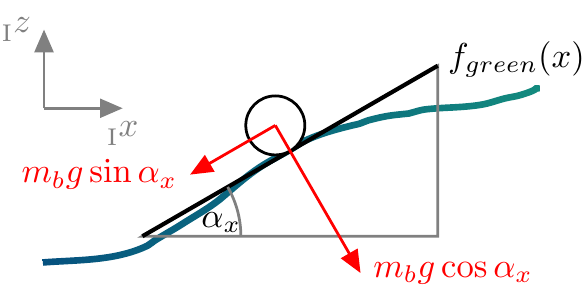}}
	\\
	\centering 
	\subfloat[rolling direction and frictional force of the ball\label{fig:inclined_plane2}]{\includegraphics[]{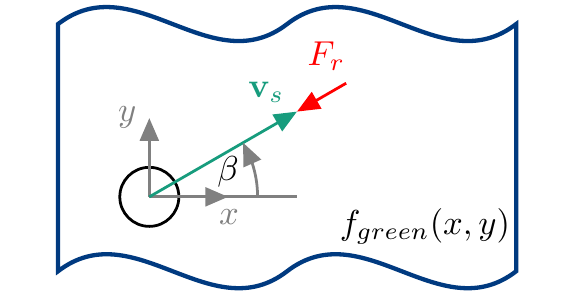}}
	\caption{The physical model of golf ball dynamics is based on inclined planes, where the frictional force $F_r$ acts parallel to the surface and opposite to the rolling direction of the ball.}
	\label{fig:inclined_plane}
\end{figure}

The surface shape of the green $f_{green}(x,y)$, cf.~Fig.~\ref{fig:inclined_plane1}, is approximated as described in Sec.~\ref{subsec:terrain_information}, so that the applied downslope forces are calculated with the angles
\begin{align}
	\alpha_x=\arctan\left(\dfrac{\partial f_{green}}{\partial x}\right), \alpha_y=\arctan\left(\dfrac{\partial f_{green}}{\partial y}\right).
\end{align}
\begin{table}[t!]
	\caption{Physical parameters of the ball dynamics.}
	\label{tab:ball_parameters}
	\begin{center}
		\begin{tabular}{c p{4.2cm} p{1.9cm}}
			%			\hline
			symbol&physical parameter&value\\
			\hline
			$m_b$& mass of the golf ball&  \SI{0.046}{\kilo\gram}\\ 
			$g$& gravity constant& \SI{9.81}{\meter\per\second^2}\\
			$\mu_b$& rolling resistance coefficient of the ball on the turf& $0.15$\\
			\hline
		\end{tabular}
	\end{center}
\end{table}
The rolling resistance
\begin{equation}
	F_r=m_bg\mu_b\cos\alpha_x \cos\alpha_y,
\end{equation}
is assumed to be constant \cite{Tra06} and always acts against the rolling direction of the ball, which is given by
\begin{equation}
	\beta=\arctan\left(\frac{\dot{y}}{\dot{x}}\right),
\end{equation}
and illustrated in Fig.~\ref{fig:inclined_plane2}.
\begin{figure*}[tb]
	\centering
	\subfloat[training of the forward neural network\label{fig:nn1}]{\includegraphics[scale=1]{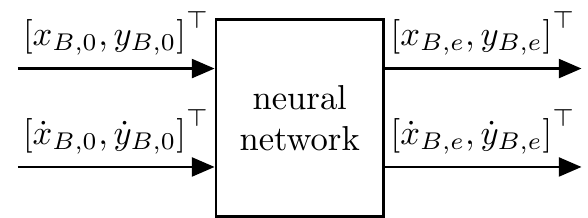}}
	%	\\
	\hspace{1cm}
	\subfloat[using the neural network with an outer optimization loop\label{fig:opt1}]{\raisebox{0.0cm}{\includegraphics[scale=1]{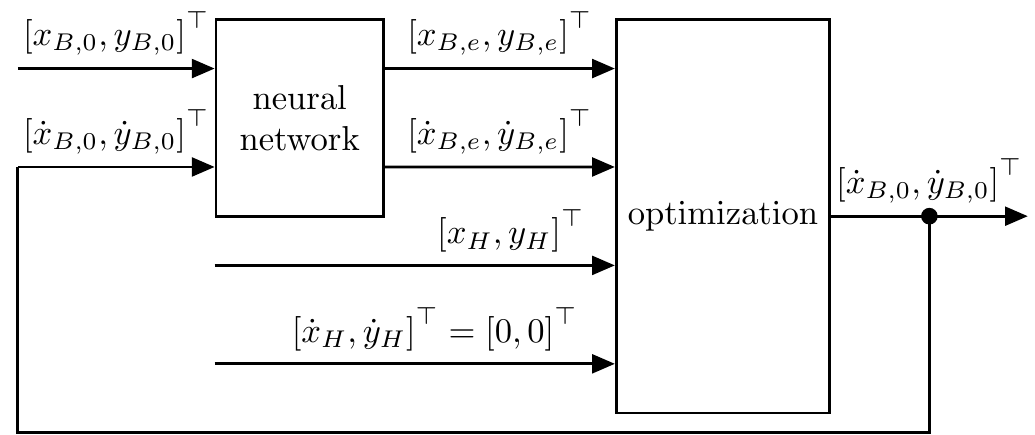}}}
	\caption{First approach: Forward predicting neural network with outer optimization loop to determine the stroke velocity vector for a given game situation. All quantities are given with respect to the inertial coordinate system I.}
\end{figure*}
\begin{figure*}[tb]
	\centering
	\subfloat[training of the inverse neural network\label{fig:nn2}]{\includegraphics[scale=1]{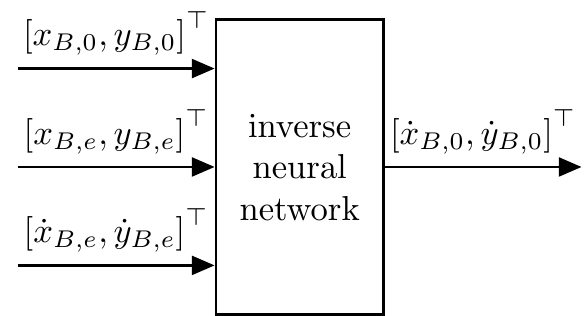}}
	%	\\
	\hspace{1cm}
	\subfloat[directly using the inverse neural network\label{fig:opt2}]{\raisebox{0.0cm}{\includegraphics[scale=1]{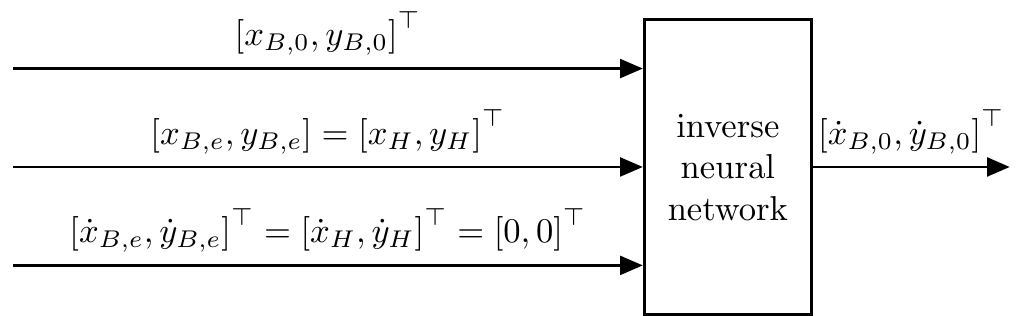}}}
	\caption{Second approach: Inverse neural network, which directly determines the required stroke velocity vector for a given game situation. All quantities are given with respect to the inertial coordinate system I.}
\end{figure*}
The resulting ball dynamics is then given by:
\begin{align}
	\begin{split}\label{eq:ball_dynamics_x}
		m_b\ddot{x}&=-m_bg\sin\alpha_x-F_r\vert\cos\beta\vert\mathrm{sgn}(\dot{x}),
	\end{split}\\
	\begin{split}\label{eq:ball_dynamics_y}
		m_b\ddot{y}&=-m_bg\sin\alpha_y-F_r\vert\sin\beta\vert\mathrm{sgn}(\dot{y}).
\end{split}
\end{align}

For pretraining, random strokes (with different initial positions and velocities) are simulated on the ball dynamics model \eqref{eq:ball_dynamics_x}-\eqref{eq:ball_dynamics_y} with $\prescript{}{\mathrm{I}}{\vec{q}}=\left[
		\prescript{}{\mathrm{I}}{x},\prescript{}{\mathrm{I}}{y},\prescript{}{\mathrm{I}}{\dot{x}},\prescript{}{\mathrm{I}}{\dot{y}}
	\right]^\top$
using RK4~solver, where the initial and end positions of the ball are given by  ${\prescript{}{\mathrm{I}}{\vec{q}_0}=\left[
	\prescript{}{\mathrm{I}}{x_0},\prescript{}{\mathrm{I}}{y_0},\prescript{}{\mathrm{I}}{\dot{x}_0},\prescript{}{\mathrm{I}}{\dot{y}_0}
\right]^\top}$ and $\prescript{}{\mathrm{I}}{\vec{q}_e}=\left[
\prescript{}{\mathrm{I}}{x_e},\prescript{}{\mathrm{I}}{y_e},0,0
\right]^\top$, respectively.
\subsection{Determine the optimal stroke velocity vector based on a neural network}\label{subsec:determine_stroke_velocity_vector}
A neural network is used to determine the optimal stroke velocity vector $\left[
	\prescript{}{\mathrm{I}}{\dot{x}_{B,0}},\prescript{}{\mathrm{I}}{\dot{y}_{B,0}}
\right]^\top$ for an initial ball position $\left[
\prescript{}{\mathrm{I}}{x_{B,0}},\prescript{}{\mathrm{I}}{y_{B,0}}
\right]^\top$, so that the ball hits the hole $\left[
\prescript{}{\mathrm{I}}{x_H},\prescript{}{\mathrm{I}}{y_H}
\right]^\top$ at zero speed, yielding
\begin{equation}
	\begin{bmatrix}
		\prescript{}{\mathrm{I}}{x_{B,e}}\\
		\prescript{}{\mathrm{I}}{y_{B,e}}
	\end{bmatrix}=\begin{bmatrix}
	\prescript{}{\mathrm{I}}{x_H}\\
	\prescript{}{\mathrm{I}}{y_H}
\end{bmatrix}, 
\begin{bmatrix}
\prescript{}{\mathrm{I}}{\dot{x}_{B,e}}\\
\prescript{}{\mathrm{I}}{\dot{y}_{B,e}}
\end{bmatrix}=\begin{bmatrix}
\prescript{}{\mathrm{I}}{\dot{x}_H}\\
\prescript{}{\mathrm{I}}{\dot{y}_H}
\end{bmatrix}=\begin{bmatrix}
0\\
0
\end{bmatrix}.
\end{equation}

Our first approach is based on a simple neural network (2~layers with 30~hidden neurons) forward predicting the golf ball dynamics, see Fig.~\ref{fig:nn1}, and a subsequent optimization loop to compute the stroke velocity vector, see Fig.~\ref{fig:opt1}. The objective function is given by 
\begin{equation}
	J_b(\prescript{}{\mathrm{I}}{\dot{x}_{B,0}},\prescript{}{\mathrm{I}}{\dot{y}_{B,0}})=(\prescript{}{\mathrm{I}}{\vec{q}_{B,e}}-\prescript{}{\mathrm{I}}{\vec{q}_{B,h}})^\top\vec{W}(\prescript{}{\mathrm{I}}{\vec{q}_{B,e}}-\prescript{}{\mathrm{I}}{\vec{q}_H}),
\end{equation} 
where $\prescript{}{\mathrm{I}}{\vec{q}_H}=\left[
	\prescript{}{\mathrm{I}}{x_H},\prescript{}{\mathrm{I}}{y_H},0,0
\right]^\top$,${\vec{W}=\text{diag}\left(1,1,1,1\right)}$ and $\prescript{}{\mathrm{I}}{\vec{q}_{B,e}}$ follows from the neural network. The optimization problem is solved using particle swarm optimization in MATLAB.

We alternatively use an inverse neural network, see Fig.~\ref{fig:nn2} (with the same number of hidden layers and neurons as the previous network), which explicitly determines the required stroke velocity vector, see Fig.~\ref{fig:opt2}, so we do not longer need to solve an optimization problem. This calculates the stroke velocity vector much faster, while the performance is the same as with the first approach. Therefore, we have solely used the inverse neural network approach in the following.
\begin{figure}[t!]
	\centering 
	\includegraphics[width=0.47\textwidth]{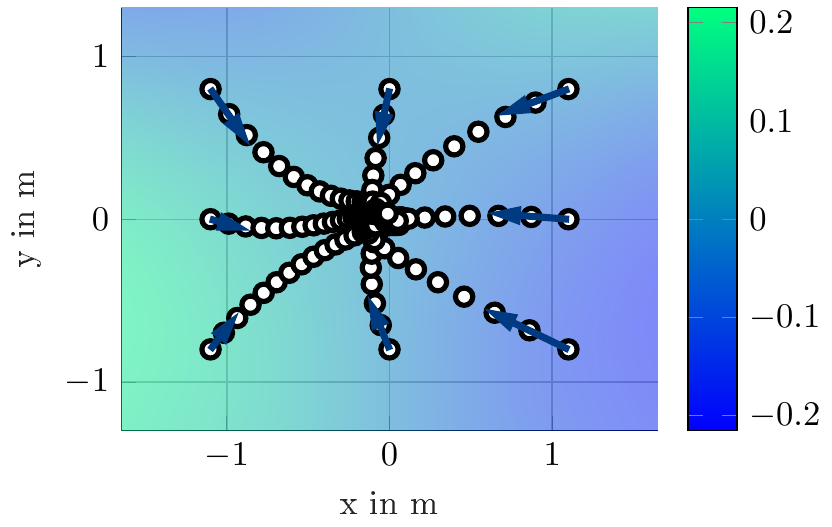}
	\caption{Model-based results of the inverse neural network for a fictional surface, where the hole is located in the origin of the coordinate system I. The initial stroke velocity vectors are visualized in blue, where the white circles mark the flow of each ball rolling trajectory over time.}
	\label{fig:worms}
\end{figure}
\begin{figure*}[t!]
	\centering
	\subfloat[initial position\label{fig:initial_position}]{\includegraphics[trim=32cm 7cm 13cm 0, clip, width=0.32\textwidth]{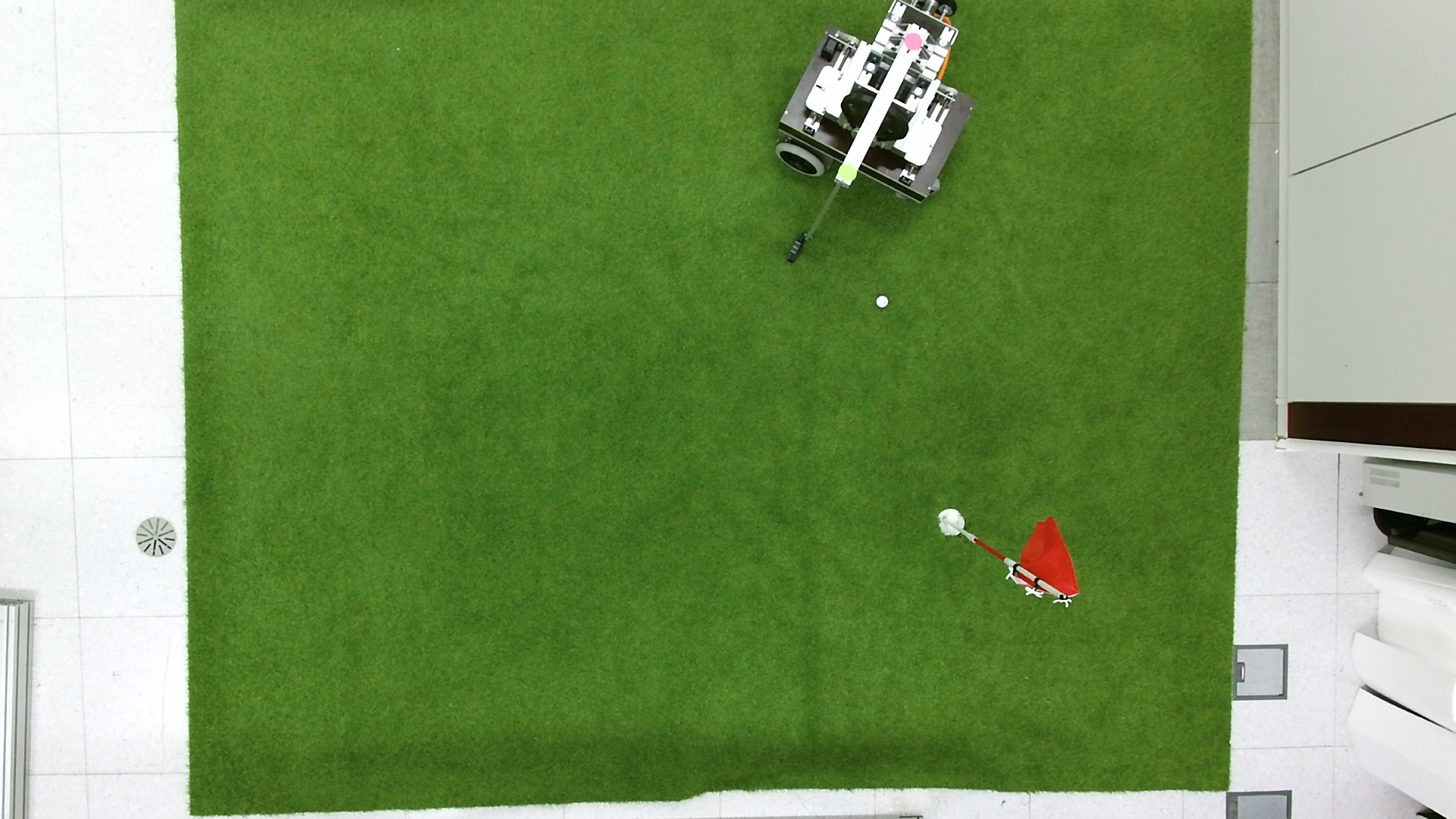}} 
	\hspace{0.01\textwidth}
	\subfloat[pre stroke\label{fig:pre_stroke}]{\includegraphics[trim=32cm 7cm 13cm 0, clip, width=0.32\textwidth]{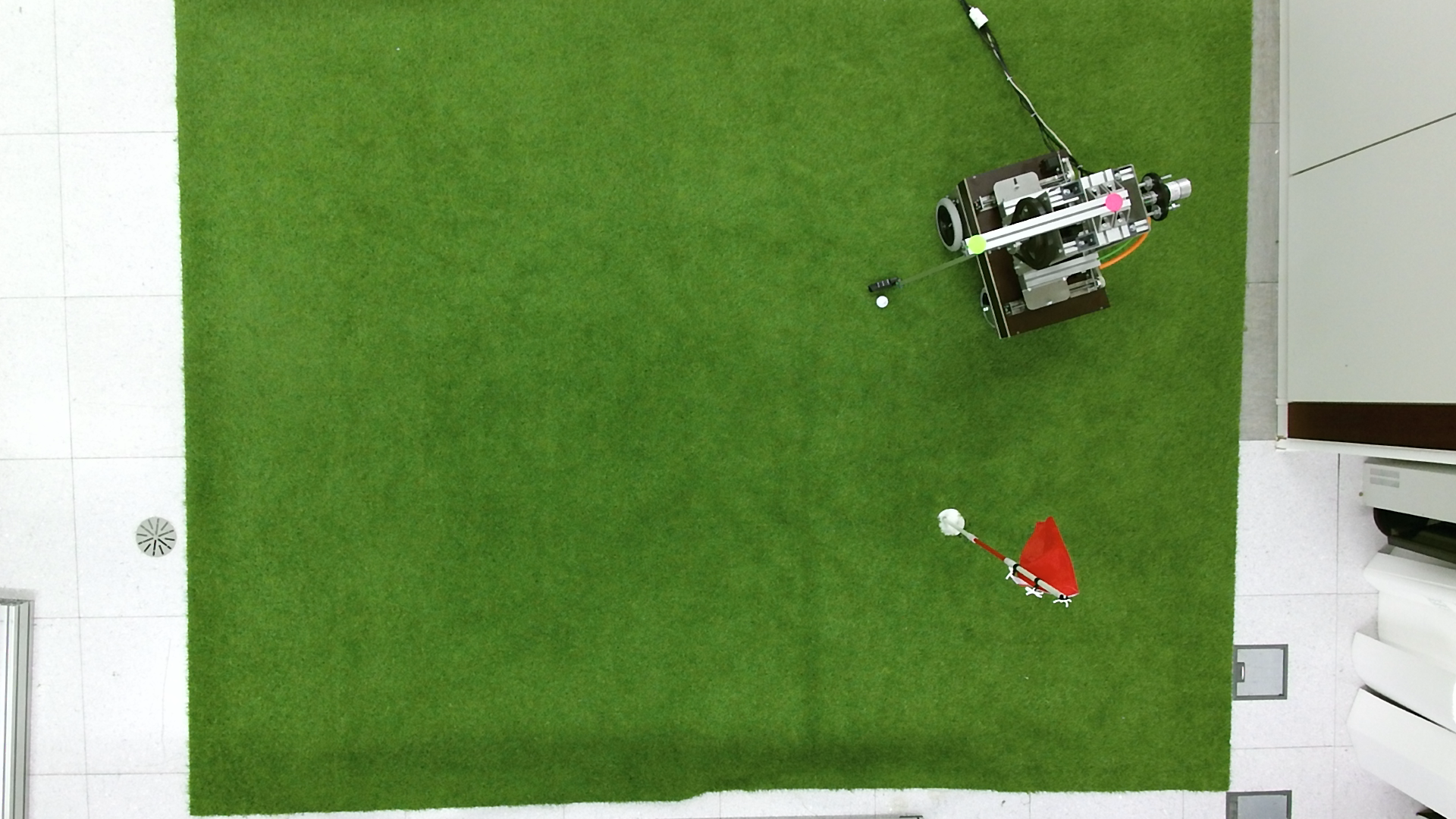}}
	\hspace{0.01\textwidth}
	\subfloat[post stroke\label{fig:post_stroke}]{\includegraphics[trim=32cm 7cm 13cm 0, clip, width=0.32\textwidth]{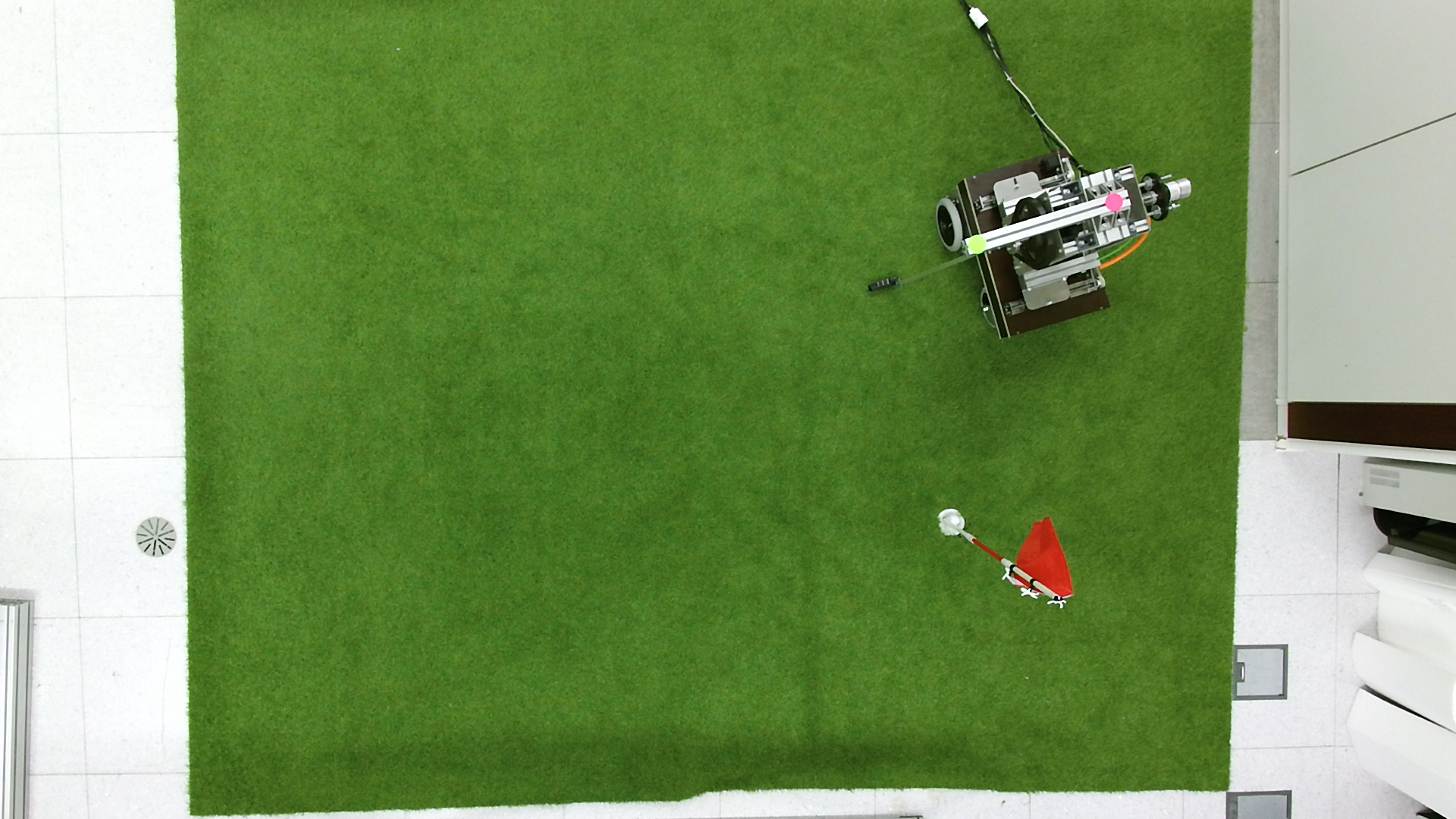}}
	\caption{Example situation and process for a golf play of the robot.}
	\label{fig:example_situation}
\end{figure*}
The model-based trained inverse neural network provides plausible results as the calculated initial stroke velocity vectors actually cause the ball to roll simulatively near or into the hole, as can be seen schematically for a fictional surface in Fig.~\ref{fig:worms}. This demonstrates the feasibility of our approach.
\section{Results}\label{sec:results}
In the following, we describe the process of an autonomous golf game with an example scenario and evaluate the success of our method. 

Before starting Golfi, the green surface is first cleared prior to being captured by the 3D camera. From this recording, the green surface is approximated and model-based training data is generated by simulating 3000~training strokes from random points at random velocities within the valid range of the green. This step is necessary only once or if the green has changed, e.g., by varying the hills. Afterwards, Golfi and the ball are placed on arbitrary initial positions. This corresponds to the situation shown in Fig.~\ref{fig:initial_position}. After that, the algorithms for the golf game (object detection, determination of the optimal stroke velocity vector, driving strategy) are executed in MATLAB. During these calculations, the user is allowed to check and confirm the detected objects as well as the proposed visualized driving trajectory. Before confirming the driving trajectory, Golfi's club is raised in a controlled manner via the dSPACE software ControlDesk to avoid collisions with the ball. The individual driving commands are now sent to Golfi via a serial interface and implemented after respective confirmation by the user, resulting in the situation shown in Fig.~\ref{fig:pre_stroke} directly before the stroke. Now the stroke is executed by transmitting and confirming the absolute value of the calculated stroke velocity vector $\lVert|\vec{v}_s\rVert|$ to ControlDesk. After the stroke has been executed, the situation shown in Fig.~\ref{fig:post_stroke} arises. If the ball hits the hole, the goal is reached. If it does not hit the hole, the process can easily be started again from the beginning. Additionally, it is feasible to use that unsuccessful stroke in combination with the detected new ball position as another training data point for retraining, as described in Sec.~\ref{sec:optimal_stroke_velocity_vector}. 

This example scenario for a surface without hills qualitatively demonstrates the success of our overall system. In practice, the hole was not always hit, but the calculations of the stroke velocity vector were plausible in all cases, so that the ball either hit the hole or stopped at a small distance from the hole. 
\section{Conclusion \& Outlook}\label{sec:conclusion}
We presented the development of an autonomous putting golf robot. First, we described the mechatronic design with the control strategy. Then we showed how Golfi intelligently determines in which direction and at what velocity it must strike the ball to make it hit the hole, using a combination of physical and data-driven methods with a 3D camera. For this, an accurate model of the ball dynamics is crucial. Without hills on the surface, Golfi performed well in moving autonomously to the ball and hitting it towards the hole.

In our work so far, we have limited ourselves to a situation where the green has no hills, but we have already demonstrated the feasibility of our method for a hilly green in a model-based manner, cf.~Fig.~\ref{fig:worms}. The next step is to evaluate these results on a hilly green in our laboratoy. For this purpose, the drive unit may no longer be sufficient and the fine traversing unit may be used as well. Further research should also aim to test and develop the pretraining strategy, cf.~Fig.~\ref{fig:pretraining_strategy}. The question will be how the real training strokes may be used for retraining, e.g., in terms of weighting compared to the model-based training strokes. In addition, it is interesting to compare the performance of Golfi with that of a human. 

\appendix[Reference trajectories for the stroke device]
\label{app}
The reference trajectories for the stroke device are mathematically described as follows
\begin{equation}
	\varphi(t)=
	\begin{cases} 
		f_1(t), & 	0<t\leq T_{l},\\
		f_2(t), &  T_{l}<t\leq \left(T_{l}+\tfrac{\varphi_{l}\pi h}{\lVert\vec{v}_s\rVert}\right),\\
		f_3(t), & 	\left(T_{l}+\tfrac{\varphi_{l}\pi h}{\lVert\vec{v}_s\rVert}\right)<t\leq \left(2T_{l}+\tfrac{\varphi_{l}\pi h}{\lVert\vec{v}_s\rVert}\right),\\
		0, & \mathrm{otherwise},
	\end{cases}\\
\end{equation}
\begin{equation}
	\dot\varphi(t)=
	\begin{cases} 
		f_4(t), & 	0<t\leq T_{l},\\
		f_5(t), &  T_{l}<t\leq \left(T_{l}+\tfrac{\varphi_{l}\pi h}{\lVert\vec{v}_s\rVert}\right),\\
		f_6(t), & 	\left(T_{l}+\tfrac{\varphi_{l}\pi h}{\lVert\vec{v}_s\rVert}\right)<t\leq \left(2T_{l}+\tfrac{\varphi_{l}\pi h}{\lVert\vec{v}_s\rVert}\right),\\
		0, & \mathrm{otherwise}
	\end{cases}
\end{equation}
with 
\begin{align}
	f_1(t)&=\dfrac{\varphi_{l}}{2}\left(\cos\left(\tfrac{\pi}{T_{l}}t\right)-1\right),\\
	f_2(t)&=-\varphi_{l}\cos\left(\tfrac{\lVert\vec{v}_s\rVert}{\varphi_{l}h}\left(t-T_{l}\right)\right),\\
	f_3(t)&=\tfrac{\varphi_{l}}{2}\left(\cos\left(\tfrac{\pi}{T_{l}}\left(t-T_{l}-\tfrac{\varphi_{l}\pi h}{\lVert\vec{v}_s\rVert}\right)\right)+1\right),\\
	f_4(t)&=-\tfrac{\varphi_{l}\pi}{2T_{l}}\sin\left(\tfrac{\pi}{T_{l}}t\right),\\
	f_5(t)&=\tfrac{\lVert\vec{v}_s\rVert}{h}\sin\left(\tfrac{\lVert\vec{v}_s\rVert}{\varphi_{l}h}(t-T_{l})\right),\\
	f_6(t)&=-\tfrac{\varphi_{l}\pi}{2T_{l}}\sin\left(\tfrac{\pi}{T_{l}}\left(t-T_{l}-\tfrac{\varphi_{l}\pi h}{\lVert\vec{v}_s\rVert}\right)\right),
\end{align}
where $\varphi_{l}$ and $T_{l}$ are the parameters for lunge and reset before and after the stroke, see Table~\ref{tab:stroke_parameters}.
\section*{Acknowledgment}
This work was developed in the junior research group DART (Daten\-ge\-trie\-be\-ne Methoden in der Regelungstechnik), Paderborn University, and funded by the Federal Ministry of Education and Research of Germany (BMBF - Bundes\-ministerium für Bildung und Forschung) under the funding code 01IS20052. The responsibility for the content of this publication lies with the authors.

We would like to thank Professor Dellnitz for the great initial idea of designing a golf robot as a demonstrator for machine learning techniques in control engineering. We also acknowledge Ricarda Götte, who inspired us with her innovative idea to use an inverse neural network to determine the stroke velocity vector avoiding the need of solving an expensive optimization problem. Thanks also to Felix Regel, who gained some first experience in the field of cameras and image processing for us. And we would especially like to emphasize our gratitude to Martin Leibenger for the creative development of all mechatronic components and first-class technical support.  

%\bibliographystyle{IEEEtran}
%\bibliography{bibliography}


\end{document}